\pdfoutput=1
\documentclass[11pt]{article}

\usepackage[]{ACL2023}
\usepackage{times}
\usepackage{latexsym}

\usepackage[T1]{fontenc}
\usepackage[utf8]{inputenc}

\usepackage{microtype}

\usepackage{inconsolata}

\usepackage{graphicx}
\usepackage{fancyvrb}

\usepackage{tabularx}
\usepackage{multirow}
\usepackage{tablefootnote}
\usepackage{booktabs}
\usepackage{amssymb}
\usepackage{amsmath}

\title{Correcting Semantic Parses with Natural Language through Dynamic Schema Encoding}

\author{Parker Glenn, Parag Pravin Dakle, Preethi Raghavan\\
  Fidelity Investments, AI Center of Excellence\\
  \{\texttt{parker.glenn, paragpravin.dakle, preethi.raghavan\}@fmr.com}}

\begin{document}
\maketitle
\begin{abstract}
In addressing the task of converting natural language to SQL queries, there are several semantic and syntactic challenges. It becomes increasingly important to understand and remedy the points of failure as the performance of semantic parsing systems improve. We explore semantic parse correction with natural language feedback, proposing a new solution built on the success of autoregressive decoders in text-to-SQL tasks. By separating the semantic and syntactic difficulties of the task, we show that the accuracy of text-to-SQL parsers can be boosted by up to 26\% with only one turn of correction with natural language. Additionally, we show that a T5-base model is capable of correcting the errors of a T5-large model in a zero-shot, cross-parser setting.
\end{abstract}

\section{Introduction}
The task of parsing natural language into structured database queries has been a long-standing benchmark in the field of semantic parsing. Success at this task allows individuals without expertise in the downstream query language to retrieve information with ease. This helps to improve data literacy, democratizing accessibility to otherwise opaque public database systems.

Many forms of semantic parsing datasets exist, such as parsing natural language to programming languages \cite{ling-etal-2016-latent, oda2015learning,quirk-etal-2015-language}, Prolog assertions for exploring a database of geographical data \cite{zelle1996learning}, or SPARQL queries for querying a large knowledge base \cite{talmor_web_2018}. The current work discusses parsing natural language into a structured query language (SQL), perhaps the most well-studied sub-field of semantic parsing.

\begin{figure}
    \centering
    \includegraphics[width=7.3cm]{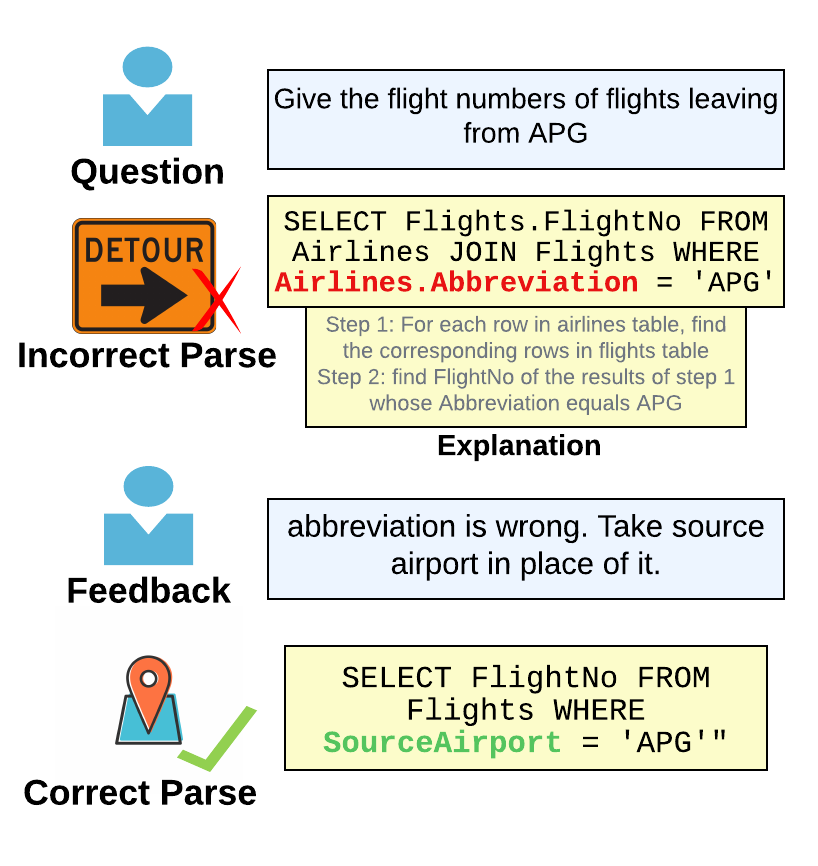}
    \caption{Example item from the \textsc{SPLASH} dataset. An incorrect parse from a neural text-to-SQL model is paired together with natural language feedback commenting on how the parse should be corrected.}
    % \caption{Example item from the \textsc{SPLASH} dataset. An incorrect parse is paired with natural language feedback }
    \label{fig:splash_example}
\end{figure}

Most text-to-SQL works frame the task as a one-shot mapping problem. Methods include transition-based parsers \cite{yin-neubig-2018-tranx}, grammar-based decoding \cite{GuoIRNet2019,lin2019grammar}, and the most popular approach as of late, sequence to sequence (seq2seq) models \cite{scholak-etal-2021-picard,qi_rasat_2022,xie-etal-2022-unifiedskg}.

In contrast to the one-shot approach, conversational text-to-SQL aims to interpret the natural language to structured representations in the context of a multi-turn dialogue \cite{yu_cosql_2019,yu-etal-2019-sparc}. It requires some form of state tracking in addition to semantic parsing to handle conversational phenomena like coreference and ellipsis \cite{zhang2019editing,hui2021dynamic,cai-etal-2022-star}. 

Interactive semantic parsing frames the task as a multi-turn interaction, but with a different objective than pure conversational text-to-SQL. As a majority of parsing mistakes that neural text-to-SQL parsers make are minor, it is often feasible for humans to suggest fixes for such mistakes using natural language feedback. Displayed in Figure \ref{fig:splash_example}, \textsc{SPLASH} (\underline{S}emantic \underline{P}arsing with \underline{L}anguage \underline{A}ssistance from \underline{H}umans) is a text-to-SQL dataset containing erroneous parses from a neural text-to-SQL system alongisde human feedback explaining how the interpretation should be corrected \cite{elgohary_speak_2020}. Most similar to \textsc{SPLASH} is the \textsc{INSPIRED} dataset \cite{mo-etal-2022-towards}, which aims to correct errors in SPARQL parses from the ComplexWebQuestions dataset \cite{talmor_web_2018}. While the interactive semantic parsing task evaluates a system's ability to incorporate human feedback, as noted in \citet{elgohary_speak_2020}, it targets a different modeling aspect than the traditional conversational paradigm. Hence, good performance on one does not guarantee good performance on the other task. 
% A parser achieving 25.16 exact match accuracy on \textsc{SPLASH} drops to 3.2 when applied to the CoSQL dataset \cite{yu_cosql_2019}.

We make the following contributions: (1) We achieve a new state-of-the-art on the interactive parsing task \textsc{SPLASH}, beating the best published correction accuracy \cite{elgohary-etal-2021-nl} by 12.33\% using DestT5 (\underline{D}ynamic \underline{E}ncoding of \underline{S}chemas using \underline{T5}); (2) We show new evidence that the decoupling of syntactic and semantic tasks improves text-to-SQL results \cite{li2022resdsql}, proposing a novel architecture which leverages a single language model for both tasks; (3) We offer a new small-scale test set for interactive parsing\footnote{\href{https://github.com/parkervg/DestT5}{https://github.com/parkervg/DestT5}}, and show that a T5-base interactive model is capable of correcting errors made by a T5-large parser. 

\section{Dataset}
In this work, we evaluate our models on the \textsc{SPLASH} dataset as introduced in \citet{elgohary_speak_2020}. It is based on Spider, a large multi-domain and cross-database dataset for text-to-SQL parsing \cite{yu2018spider}. Incorrect SQL parses were selected from the output of a Seq2Struct model trained on Spider \cite{shin_encoding_2019}. Seq2Struct achieves an exact set match accuracy of 42.94\% on the development set of Spider.

Alongside the incorrect parse, an explanation of the SQL query is generated using a rule-based template. Annotators were then shown the original question $q$ alongside the explanation and asked to provide natural language feedback $f$ such that the incorrect parse $p'$ could be resolved to the final gold parse $p$. 

Each item in the \textsc{SPLASH} dataset is associated with a relational database $\mathcal{D}$. Each database has a schema $\mathcal{S}$ containing tables $\mathcal{T} = \{t_1, t_2, ... t_N\}$ and columns $\mathcal{C} = \{c^1_{1},...,c^1_{n_1}, c^2_{1},...,c^2_{n_2}, c^N_{1},...,c^N_{n_N}\}$, where $N$ is the number of tables, and $n_i$ is the number of columns in the $i$-th table. Figure \ref{fig:splash_example} displays an example item from the \textsc{SPLASH} dataset, excluding the full database schema $S$ for brevity. 
\defcitealias{yu2018spider}{Tao Yu et al. (2018)}
\section{Model}

\begin{figure*}
    \centering
    \includegraphics[width=\textwidth]{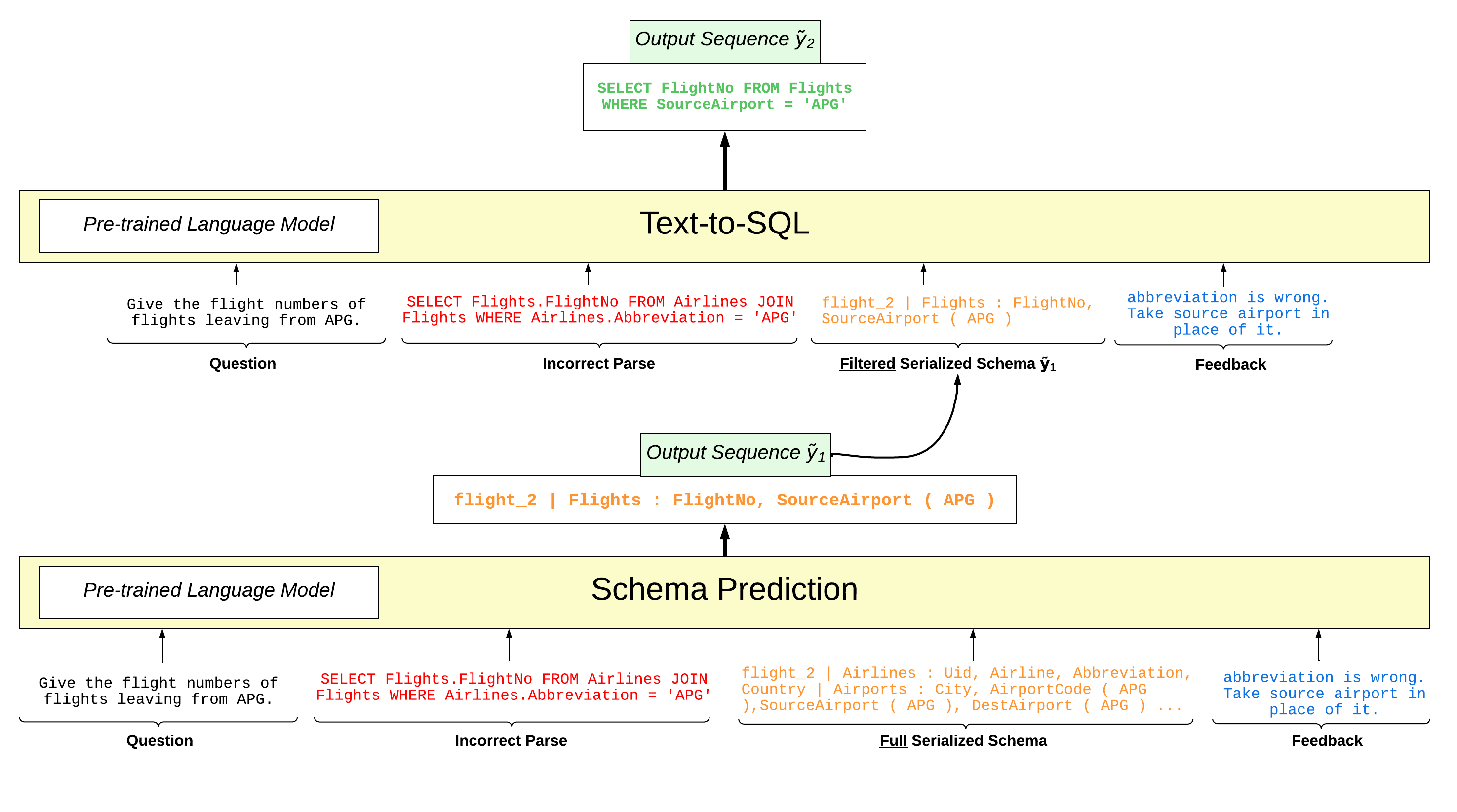}
    \caption{Model architecture. In ``Schema Prediction'', the database schema is filtered to only the relevant items $\tilde{y}_1$ using a classifier or generator described in Section \ref{dynamic_schema_encoder}. In ``Text-to-SQL'', the output of the schema prediction model is used to generate the final parse $\tilde{y}_2$.}
    \label{fig:model_architecture}
\end{figure*}

\subsection{Dynamic Schema Encoder}\label{dynamic_schema_encoder}
In converting natural language to SQL, a parser must handle both the semantic challenges in selecting the correct tables and columns from the database schema, and generate valid SQL syntax. As shown in \citet{li2022resdsql}, decoupling the schema linking and skeleton parsing tasks in text-to-SQL improves results when applied to the Spider dataset. We take a similar approach with the \textsc{SPLASH} dataset, separating the semantic and syntactic challenges of text-to-SQL by introducing an auxiliary schema prediction model. This auxiliary model serializes only the most relevant schema items into the input for the final seq2seq text-to-SQL model. 

The task of the schema prediction is to output only those schema items (tables, columns, values) that appear in the gold SQL $p$. The inputs can be represented as follows.

\begin{equation}
\begin{aligned}
\label{eq:serialized_schema}
d = t_1 : c^1_{1},...,c^1_{n_1} |...|t_N : c^N_{1},...,c^N_{n_N}
\end{aligned}
\end{equation}
\begin{equation}
\begin{aligned}
\label{eq:schema_inputs}
x = ([CLS],q,[SEP],d,[SEP],p',[SEP],f)
\end{aligned}
\end{equation}

Where $d$ represents a flattened representation of the database schema $\mathcal{S}$, $q$ is the question, $p'$ is the incorrect parse from \textsc{SPLASH}, and $f$ is the natural language feedback. For each schema item, the task is to predict the presence or absence of the item in the final gold SQL parse $p$.

By introducing this auxiliary schema prediction model, the final text-to-SQL model should only be tasked with stitching together the predicted schema items into valid SQL logic. As shown in the example in Figure \ref{fig:model_architecture}, the text-to-SQL model is able to filter out the unnecessary ``join'' clauses from the incorrect parse, given the only table predicted by the schema prediction is ``Flights''. 

This approach was validated by carrying out a simple experiment. We serialize only those ``gold'' schema items that appear in the translated SQL and fine-tune a T5-base model\footnote{\href{https://huggingface.co/tscholak/t5.1.1.lm100k.base}{https://huggingface.co/tscholak/t5.1.1.lm100k.base}} on the Spider dataset to achieve a best 78.10\% execution accuracy. This beats the vanilla T5-base model\footnote{\href{https://huggingface.co/tscholak/1zha5ono}{https://huggingface.co/tscholak/1zha5ono}} by 18.7\%, demonstrating that successful schema prediction sets up a text-to-SQL model to predict the final query with high accuracy. 

% By alleviating the semantic difficulties, a text-to-SQL parser is highly successful in inserting the schema items into correct SQL syntax.

% by alleviating some of the semantic difficulties.

\paragraph{Schema Classifier}
We adopt the RoBERTa-large schema prediction described in \citet{li2022resdsql} for our classification model. To alleviate the label imbalance problem caused by sparse schema targets, focal loss is used as the loss function \cite{lin2017focal}. Focal loss adds a factor $(1 - p_t)^\gamma$ to standard cross entropy loss, reducing relative loss for well-classified examples and putting more focus on misclassified examples.

\begin{equation}
\begin{aligned}
    &\mathcal{L}_2 = \frac{1}{N}\sum_{i=1}^{N} FL(y_{i}, \hat{y}_i) +  \frac{1}{M}\sum_{i=1}^{N} \sum_{k=1}^{n_i} FL( y^{i}_{k}, \hat{y}^i_{k})\\
\end{aligned}
\end{equation}

Where $FL$ denotes the focal loss function. $y_i$ is the ground truth label of the $i$-th table, either 0 or 1 indicating the presence or absence, respectively. Similarly, $y^{i}_{k}$ is the ground truth label of the $k$-th column in the $i$-th table.

Rather than using a hard probability threshold, hyperparameters $k_1$ and $k_2$ are introduced. Taking the probabilities from the cross-encoder, only the top-$k_1$ tables and top-$k_2$ columns are kept and serialized into a ranked schema serialization, descending by probability.

\paragraph{Schema Generator}
In addition to the previously discussed RoBERTa-large cross-encoder, we also experiment with a generative schema prediction model. T5 (Text-to-Text Transfer Transformer) is a transformer-based encoder-decoder model that converts all NLP problems into a text-to-text format \cite{raffel2020exploring}. In our task setup, the encoder applies its bidirectional attention mechanism over the features from \textsc{SPLASH} and the serialized schema items, depicted in Equation \ref{eq:schema_inputs}. The decoder, then, generates the correct SQL parse, employing teacher forcing during the training phase. It is fine-tuned using standard cross-entropy loss.

\begin{equation}
\begin{aligned}
&\mathcal{L}_1=-\sum_{i=1}^My_i\log(\hat{y_i})
\end{aligned}
\end{equation}

The target label $y_i$ will always take the form of tokens comprising the gold schema items, i.e., those tables and columns that appear in the correct SQL parse. We format the multi-label targets $y$ as text following the structure shown below. Note that this is the same structure we use to serialize the flattened database schema $d$ in Equation \ref{eq:serialized_schema}.
\begin{Verbatim}
  [db_id] | [table] : [column] (...)
\end{Verbatim}

As the theoretical output space of $\hat{y}$ is the unconstrained vocabulary of the T5 model, schema hallucinations are possible, and column/table pairs may be generated that do not exist in the database context\footnote{We note that \citet{scholak-etal-2021-picard} offers a solution for these schema hallucinations, but leave the integration of Picard to future work.}. A trade-off in this approach, however, is that the generation objective allows us to bypass the need for hyperparameters $k_1$ and $k_2$, as we simply keep the greedy argmax of $\hat{y}$ directly at each timestep. As shown in Table \ref{schema_prediction_results}, this optimization objective results in far greater precision than the classification approach but suffers a drop in recall.

\subsection{Text-to-SQL Encoder/Decoder}
We use a T5-base model to encode the unified input (with schema predictions) and generate the SQL query \cite{raffel2020exploring}.

\subsection{SQL Normalization}
We follow the same normalization procedure described in \citet{li2022resdsql}. Specifically, we normalize both the incorrect parses and gold SQL queries by (1) replacing table aliases with their original names, (2) adding an \textit{ASC} keyword if \textit{ORDER BY} doesn't already specify, (3) lower-casing all text, and (4) adding spaces around parentheses and replacing double quotes with single quotes.

\begin{table}
\begin{tabularx}{\columnwidth}{l|lll}
\hline
\textbf{Schema Model} & \textbf{F1} & \textbf{Precision} & \textbf{Recall} \\ \hline
Generator                         & \textbf{88.98}       & 90.84               & 89.18            \\
Classifier                    & 34.50       & 22.12              & \textbf{94.41}           \\\hline

\end{tabularx}
\caption{Performance of schema prediction models in predicting gold schema items on the \textsc{SPLASH} test set. Note that the classification-based method of \citet{li2022resdsql} trades low precision for high recall\tablefootnote{Not considered in this table is the ranking-enhanced nature of the RoBERTa-large method.}.}
\label{schema_prediction_results}
\end{table}

\begin{table*}[]
\centering
\begin{tabular}{@{}lll|ll@{}}
\toprule
                                                    &                       &                  & \multicolumn{2}{l}{\textbf{Shuffled Feature EM\% Change}} \\ \midrule
                                                    & \textbf{Schema Model} & \textbf{EM\%} & \textbf{Feedback}     & \textbf{Incorrect Parse}     \\ \midrule
\multicolumn{1}{l|}{\multirow{3}{*}{All}}           & None                  & 41.17            &   -                    &   -                           \\
\multicolumn{1}{l|}{}                               & Generator             & 51.35            & -2.17                 & -28.27                       \\
\multicolumn{1}{l|}{}                               & Classifier            & 49.79            & -2.7                  & -11.64                       \\ \midrule
\multicolumn{1}{l|}{\multirow{2}{*}{- Question}}    & Generator             & 48.96            & -4.47                 & -30.77                       \\
\multicolumn{1}{l|}{}                               & Classifier            & 35.97            & -11.23                & -29.94                       \\ \midrule
\multicolumn{1}{l|}{\multirow{2}{*}{- Explanation}} & \textbf{Generator}             & \textbf{53.43}            & -1.77                 & -18.09                      \\
\multicolumn{1}{l|}{}                               & Classifier            & 49.27            & -2.08                 & -17.57                       \\ \midrule
\multicolumn{1}{l|}{\multirow{2}{*}{\begin{tabular}[c]{@{}l@{}}- Question\\ - Explanation\end{tabular}}} & Generator & 47.00 & -5.53 & -38.68 \\
\multicolumn{1}{l|}{}                               & Classifier            & 38.98            & -12.47                & -36.9                        \\ \bottomrule
\end{tabular}
\caption{Results on \textsc{SPLASH} test set with various features and schema prediction models. \textit{Generator} refers to the T5-large model, and \textit{Classifier} refers to the RoBERTa-large model of \citet{li2022resdsql}. The models are evaluated on the test set with shuffled features to examine the extent to which they utilize the unique interactive components of the parsing task. In bold is DestT5.}
\label{model_ablations}
\end{table*}

% \begin{table*}
% \centering
% \begin{tabular}{lllll}
% \hline
%                       & \textbf{Correction Acc. (\%)}   & \textbf{Edit ↓ (\%)} & \textbf{Edit ↑ (\%)} & \textbf{Progress (\%)} \\ \hline
% EditSQL+Feedback      & \multicolumn{1}{l|}{25.16} & 47.44           & 23.51           & 7.71              \\
% NL-Edit               & \multicolumn{1}{l|}{41.17} & 72.41           & 16.93           & 36.99             \\ \hline
% DestT5 (Ours)          & \multicolumn{1}{l|}{\textbf{53.43}} & \textbf{75.36} & \textbf{10.38} & \textbf{56.29} \\
%  \hline
% \end{tabular}
% \caption{Comparing DestT5 to the \textsc{SPLASH} baselines EditSQL + Feedback \cite{elgohary_speak_2020} and NL-EDIT \cite{elgohary-etal-2021-nl}. }
% \label{edit_changes}
% \end{table*}

% \begin{figure}
%     \centering
%     \includegraphics[width=7cm]{img/t5_large_error_rate_spider.png}
%     \caption{Error rates on the Spider dev split, using a fine-tuned T5-large. This model achieved an exact match score of 71.2 on the dev split.}
%     \label{fig:t5_large_error_rates}
% \end{figure}

\section{Experiments}

\begin{figure}
    \centering
    \includegraphics[width=7cm]{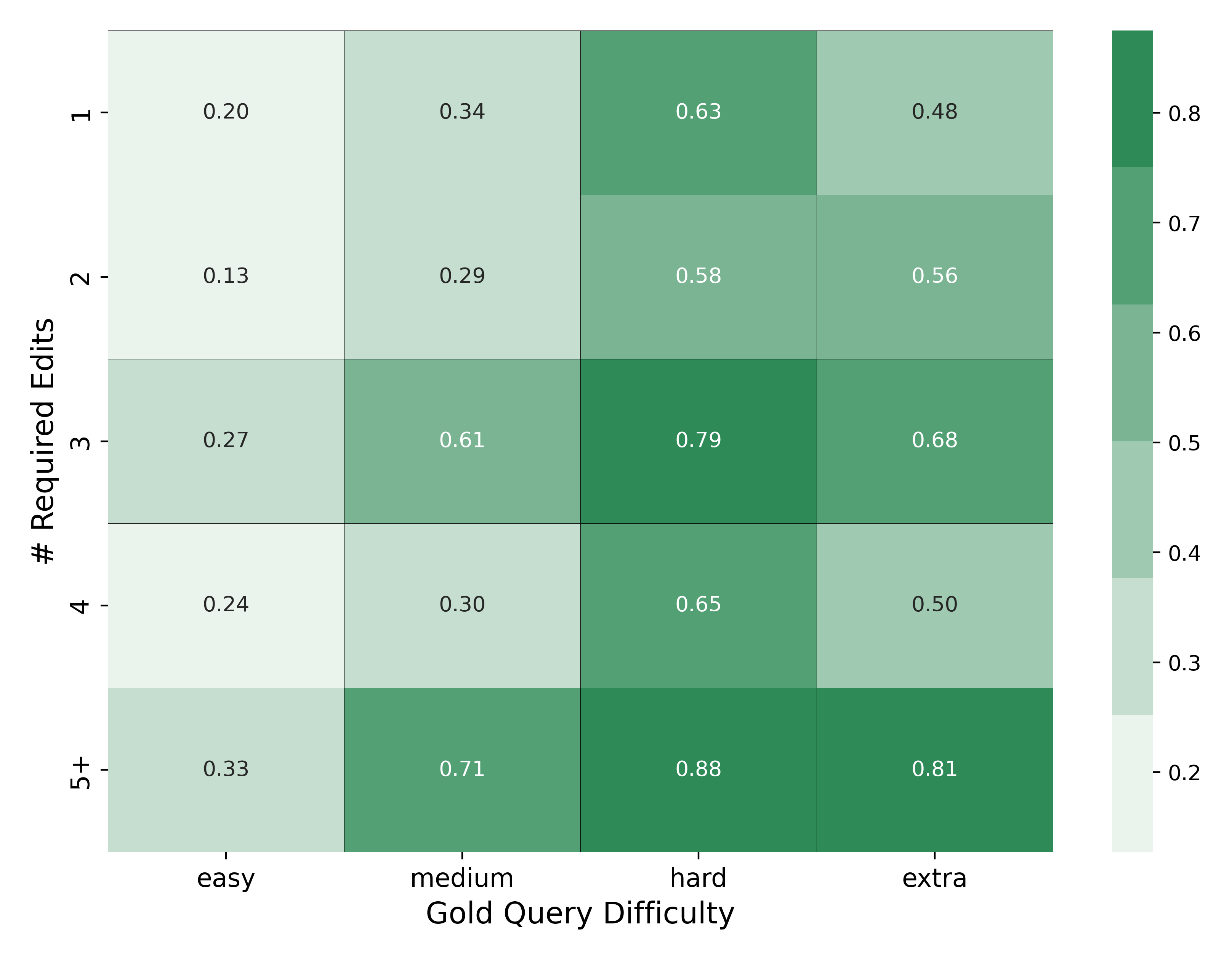}
    \caption{DestT5 error rates on the \textsc{SPLASH} test set, using the Spider exact match metric. As the distance (\textit{ \# Required Edits}) from the incorrect parse to the gold query increases, error rates also increase.}
    \label{fig:error_rates_heatmap}
\end{figure}
\subsection{Experimental Setup}

We run a series of experiments on the \textsc{SPLASH} dataset to evaluate the robustness of the proposed method. The training set contains 2,775 unique questions from the train split of Spider. \textsc{SPLASH} annotators were also asked to generate paraphrases for a single piece of feedback to improve diversity, resulting in a total of 7,481 items in the train split. The \textsc{SPLASH} test set is based on 506 items from the Spider dev split, coming out to 962 total test items with paraphrasing. 

\subsection{Evaluation Metrics}

\paragraph{Exact Set Match (EM)}
This metric evaluates the structural correctness of the predicted SQL. It checks for an orderless set match between each component in the predicted and gold query, ignoring predicted values. Many early text-to-SQL models only report EM accuracy.

\paragraph{Execution Accuracy (EX)}
Execution accuracy compares the execution results of the predicted SQL query and the gold SQL query. Since two SQL queries that do not have an exact set match may execute to the same results (e.g. ``\verb|...ORDER BY val ASC LIMIT 1|'' and ``\verb|SELECT MAX(val)|''), this metric serves as a performance upper bound. However, this metric can suffer from a high false positive rate. For this reason, we use the test suite execution accuracy with optimized database values described in \citet{zhong-etal-2020-semantic}.

\subsection{Implementation Details}

\paragraph{Text-to-SQL}
All text-to-SQL models use a fine-tuned T5-base. We use the same hyperparameters specified in the \textsc{PICARD} codebase\footnote{\url{https://github.com/ServiceNow/picard}}. Models were fine-tuned with Adafactor \cite{shazeer2018adafactor} with a learning rate 1e-4, batch size 16 for 256 epochs. A linear warm-up for the first 10\% of training steps is employed, followed by cosine decay.

\paragraph{Schema Generation}
T5-large was used for the schema generation model. It was fine-tuned using Adafactor with a constant learning rate of 1e-4 and a batch size of 4 for 512 epochs. 

\paragraph{Schema Classification}
For the schema classification model, we follow the implementation and hyperparameters described in \citet{li2022resdsql}. Specifically, we train a cross-encoder based on RoBERTa-large \cite{liu2019roberta}. AdamW \cite{loshchilov2017decoupled} with a batch size of 32 and a learning rate of 1e-5 is used for optimization. Focal loss is used to alleviate the label-imbalance problem that comes from sparse schema targets. The threshold hyperparameters $k_1$ and $k_2$ are set to 4 and 5, respectively. Specifically, only the top-4 tables and top-5 columns with the highest logits are kept and serialized as a ranked input to the text-to-SQL model.

\begin{table*}[]
\begin{tabular}{@{}llllll@{}}
\toprule
                                               & \textbf{Seq2Struct (\textbf{SPLASH})} & \textbf{EditSQL} & \textbf{TaBERT}  & \textbf{RAT-SQL} & \textbf{T5-Large} \\ \midrule
Spider Dev EM\%                                & 41.3                & 57.6    & 65.2    & 69.7    & 71.2     \\
Spider Dev EX\%                                & -             & - & - & -     & 74.4     \\ \midrule
                                               &                     &         &         &         &          \\ \midrule
\large{\textsc{\textbf{NL-EDIT}}}                              &                     &         &         &         &          \\ 
\textsc{SPLASH} Test Set EM\%     & 41.1                & 28      & 22.7    & 21.3    & -  \\
\textsc{SPLASH} Test Set EX\% & - & - & - & - & - \\
EM $\Delta$ w/ Interaction                     & +20.3               & +8.9    & +5.9    & +4.3    & -  \\
EX $\Delta$ w/ Interaction                     & -             & - & - & - & -  \\ \midrule
                                               &                     &         &         &         &          \\ \midrule
\large{\textsc{\textbf{DestT5 (Ours)}}}                                &                     &         &         &         &          \\ 
\textsc{SPLASH} Test Set EM\% & 53.43               & 31.82   & 31.47   & 28.37   & 26.1     \\
\textsc{SPLASH} Test Set EX\% & 56.86               & 40.3    & 28.84   & 36.53   & 30.43    \\
EM $\Delta$ w/ Interaction                     & \textbf{+26.15}              & \textbf{+10.16}  & \textbf{+8.13}   & \textbf{+5.71}   & \textbf{+2.83}    \\
EX $\Delta$ w/ Interaction                     & -             & - & - & -   & +3.3     \\ \bottomrule
\end{tabular}
\caption{Evaluating zero-shot generalization of DestT5 to other modern parsers. Shown are the scores without interaction on the full Spider dev set, as well as the \textbf{$\Delta$ w/ Interaction} on the Spider dev set following single-turn corrections with \textsc{NL-EDIT} and \textsc{DestT5}. This change is a byproduct of the size of the test sets (962, 330, 267, 208, and 112 left-to-right), and it is expected to increase proportional to the reported \textbf{Test Set EM\%/EX\%} as the size of the dataset increases. We indicate instances where the scores are not publicly available for a given model with \textbf{-}.}
\label{all_test_sets}
\end{table*}

\begin{figure*}[ht!]
    \centering
    \includegraphics[width=12cm]{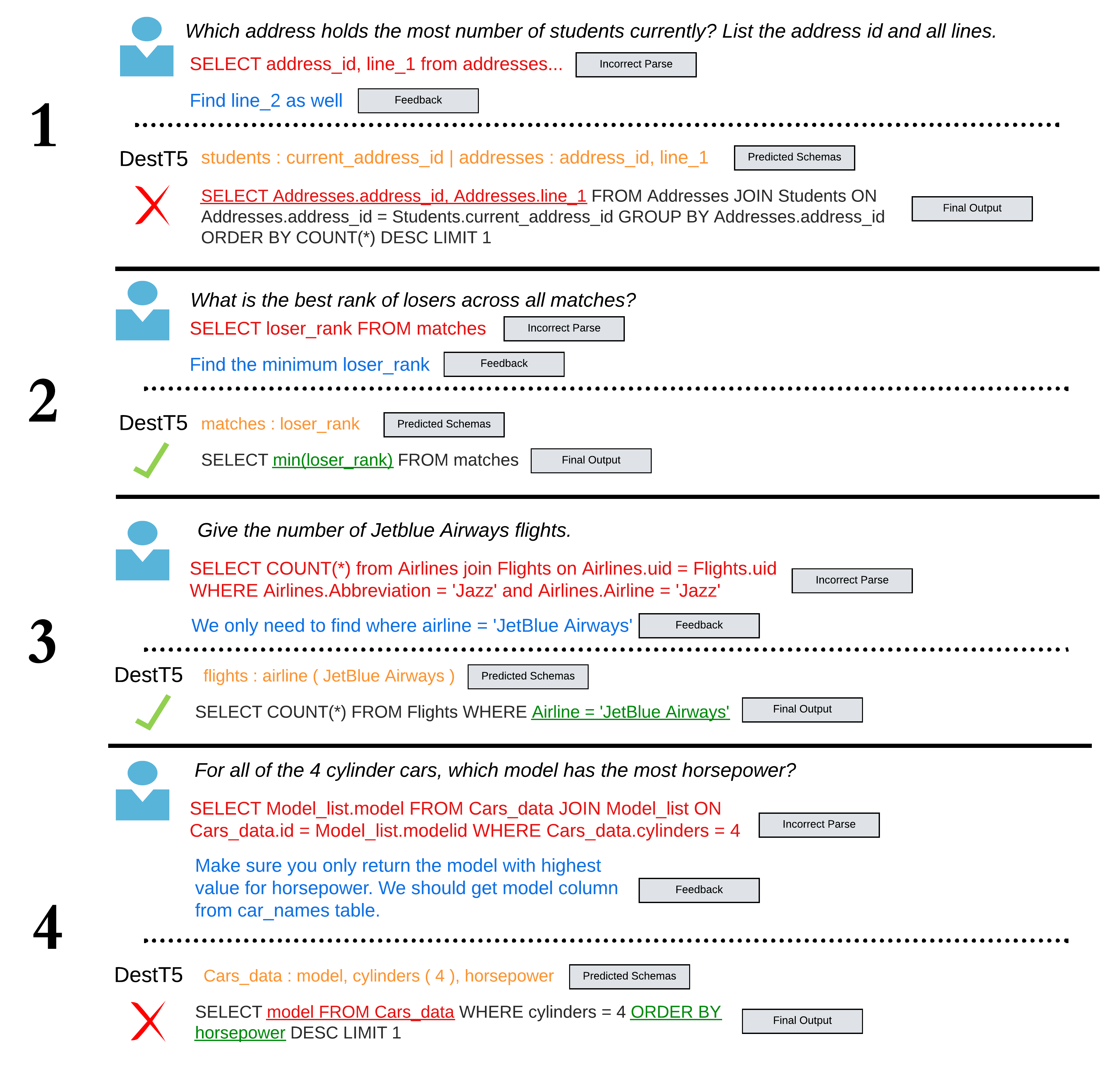}
    \caption{Example outputs of DestT5 on errors made with a T5-large text-to-SQL model. When the schema prediction model fails to identify schema items, the final text-to-SQL output is incorrect. However, when the schema prediction model is correct, it allows the text-to-SQL component to focus its efforts on generating valid SQL syntax, faithful to the feedback. See section \ref{error_analysis} for more detailed analysis of these examples.}
    \label{example_outputs}
\end{figure*}
\subsection{Evaluation}
Unlike the Spider dataset, performance on the \textsc{SPLASH} dataset is more nuanced and must be viewed holistically. To this end, we plot both ``Exact Match \%'' and ``Shuffled Feature Change'' in Table \ref{model_ablations}. The ideal model is one that achieves a competitive exact match metric, while experiencing a large drop in performance with shuffled feedback and incorrect parses\footnote{We note that a T5-base model fine-tuned with the Spider train set achieves 50.00 EM on the \textsc{SPLASH} test set.}. We find the highest exact match accuracy when removing the explanation of the incorrect parse, and by using a T5-based generative schema prediction model. This model, denoted in bold in Table \ref{model_ablations}, is later referred to as DestT5 (\underline{D}ynamic \underline{E}ncoding of \underline{S}chemas using \underline{T5}). Achieving an EM score of \textbf{53.43\%}, DestT5 beats the previous best score of NL-EDIT by 12.33\% \cite{elgohary-etal-2021-nl}. 

Using the scripts provided from \citet{elgohary-etal-2021-nl} to count SQL edits, we plot error rates on the \textsc{SPLASH} test set for both gold query difficulty and the number of edits. ``Difficulty'' is defined by \citet{yu2018spider} and classifies each SQL query into one of four categories depending on the complexity of the query. As seen in the heatmap, error rates share a positive correlation with both SQL difficulty and \# edits required to reach the gold parse. 
\begin{table*}[ht!] 
\centering
\begin{tabular}{lll}
\hline
\textbf{Text-to-SQL Model}                  & \textbf{Schema F1} & \textbf{\# Hallucinated Schema Items} \\ \hline
T5-large\tablefootnote{\href{https://huggingface.co/tscholak/3vnuv1vf}{https://huggingface.co/tscholak/3vnuv1vf}} & 79.00    & 92  \\
T5-base  & 73.92 & 121 \\
DestT5   & \textbf{80.09}  & 59                                    \\ \hline
\end{tabular}
\caption{Analysis of the schema items produced by the final text-to-SQL model. DestT5, with an auxiliary schema prediction model, identifies the presence of gold schema items with a higher F1 than a T5-large text-to-SQL model alone.}
\label{text2sql_schema_scores}
\end{table*}

\subsection{Generalizing to Other Parsers}

In recent years, massive strides have been made in the task of semantic parsing. Since the release of the \textsc{SPLASH} dataset, variations of T5 have largely taken the top spots in the Spider leaderboard. As of April 2023, all 6 models in the top 10 with corresponding publications build off of some T5 model. It is fair, then, to ask if performance on the \textsc{SPLASH} dataset actually corresponds to the ability to fix errors made with modern parsing systems, such as those utilizing T5. 

To this end, we evaluate DestT5 on the crowdsourced test sets\footnote{\href{https://github.com/MSR-LIT/NLEdit}{https://github.com/MSR-LIT/NLEdit}} based on errors made by EditSQL \cite{zhang2019editing}, TaBERT \cite{yin-etal-2020-tabert}, and RAT-SQL \cite{wang-etal-2020-rat}. Additionally, we compile a new, small-scale test set of errors made by a fine-tuned T5-large model\footnote{\href{https://huggingface.co/tscholak/3vnuv1vf}{https://huggingface.co/tscholak/3vnuv1vf}} on the Spider dev set. It contains 112 items annotated with feedback referencing the erroneous parse made by the model and is later referred to as the ``T5-large Test Set''.

Table \ref{all_test_sets} plots the end-to-end accuracy of DestT5. As mentioned in \citet{elgohary-etal-2021-nl}, there is a notable drop in the end-to-end gains as the accuracy of the base parser improves. This is likely due to the fact that as parsers improve, most of the errors are based on very complex gold SQL queries.
% We visualize this by plotting the error rates of the T5-large model in Figure \ref{fig:t5_large_error_rates}.

% As seen in Table \ref{model_ablations}, the question is an important signal for both the schema prediction and text-to-SQL task. Without the question present when training the auxiliary schema prediction model, the end-to-end exact match score with the generative model drops from 47 to TODO. [Describe how in simple queries, just the schema encoding can guide to the correct feedback without explicit feedback]

\subsection{Error Analysis}
% \textcolor{red}{TODO: add breakdown of error types (schema, skeleton, etc.)} 

% \textcolor{red}{TODO: Demonstrate how the generate schema encoding is more robust to value discrepencies (e.g. the 'Carribean', 'caribean' distinction} 
\subsection{Errors on T5-Large Test Set}\label{error_analysis}
Figure \ref{example_outputs} depicts the outputs of a randomly selected set of interactions from the T5-large test set. We discuss some of the examples below.

In Example 1, the original T5-large text-to-SQL model fails to map the phrase ``all lines'' to both columns \textit{line\_1} and \textit{line\_2}. However, even with the feedback ``Find line\_2 as well'', the auxiliary schema prediction model fails to select ``line\_2'' as a schema candidate. As a result, the final DestT5 text-to-SQL is not equipped with enough context to generate the correct parse.

In Example 2, an `easy' gold query (``\verb|SELECT MIN(loser_rank) FROM matches|'') is incorrectly parsed. This is likely due to the same reason described in \citet{lin_bridging_2020}, characterized by difficulty in mapping ``predominantly'' to \textit{spoken by the largest percentage of the population}: it remains challenging for large pre-trained models to ground terms like ``best rank'' to the DB schema. Pre-training tasks have been proposed in attempts to further improve schema grounding in LLMs, but more work can be done to align LLMs with lexical constructs grounded to the syntax of semantic parsing tasks \cite{deng_structure-grounded_2021,yin-etal-2020-tabert}. In one turn of interaction with DestT5, this syntax error is corrected.

Example 4 displays an interaction parsing long feedback with mixed success. The interaction allows DestT5 to remedy the missed semantic mapping from ``most horsepower'' to the ``\verb|ORDER BY horsepower|'' clause, but it hallucinates the ``\verb|Cars_data|'' from the ``\verb|model|'' table, failing to learn from the feedback saying otherwise. 

\section{Discussion}

% \begin{figure}
%     \centering
%     \includegraphics[width=7cm]{img/interactive_evaluation_difficulty.png}
%     \caption{Venn diagram highlighting difficulty of evaluating interactive parsing models. As ``Model A'' corrects all items that the base model already parses correctly, ``Model B'' would be the preferred model, despite the lower Exact Match (EM) performance.}
%     \label{fig:interactive_evaluation_difficulty}
% \end{figure}

\subsection{Impact of Auxiliary Schema Prediction}
Table \ref{model_ablations} displays the EM of a standard text-to-SQL model with no auxiliary schema prediction (with all schema items directly serialized as input). As shown, the score drops from 51.35\% with an auxiliary generator to 41.17\% without. We hypothesize that given the increased number of features in interactive semantic parsing (explanation, feedback, incorrect parse), distilling the role of the text-to-SQL model to primarily handling syntax parsing prevents excessive proliferation of feature interactions. 

Table \ref{text2sql_schema_scores} displays the schema F1 scores of various text-to-SQL models. Schema F1 is calculated by comparing those schema items (tables, columns) generated in the predicted parse to the schema items in the gold SQL. As shown, implementing a dedicated schema prediction model into a text-to-SQL pipeline helps identify those gold schema items with a higher F1 score, and minimizes schema hallucinations (i.e. generating tables/columns not present in the database schema). 

\paragraph{How often does the text-to-SQL model use the predicted schemas?}
We evaluate the usage rates of the predicted schema items by the final text-to-SQL model. Specifically, we examine the rate at which DestT5 either predicts a schema item not directly serialized by the schema prediction model, or fails to integrate a schema item that was serialized. We find that on the \textsc{SPLASH} test set, there are 112 instances of overpredictions by the text-to-SQL model and 210 underpredictions. There is an average distance of 0.81 between the serialized schema items and gold schema items, and 0.93 between the schema items predicted by the text-to-SQL model and gold. This indicates that, if the text-to-SQL model were explicitly restricted to use only the schema items generated by the auxiliary schema prediction model, performance will improve. We leave this and other combinations of the two models (such as joint training) to future work.

\subsection{Evaluating Interactive Parsing}

The goal of interactive semantic parsing is not to parse the most interactions correctly on the \textsc{SPLASH} test set, but more specifically to parse those interactions correctly that the original text-to-SQL model parsed incorrectly. For example, if a hypothetical interactive parsing model $A$ achieves a high EM\% on the \textsc{SPLASH} test set, but the ``$\Delta$ w/ Interaction'' metric with modern parsers is small, then the model serves minimal utility in an actual conversational setting. On the other hand, if a model $B$ performs poorly on the \textsc{SPLASH} test set but demonstrates a high ``$\Delta$ w/ Interaction'', we would deem this model as the better interactive semantic parser.

We argue, then, that the ``Correction Acc. (\%)'' metric from \textsc{SPLASH} should be replaced in favor of the end-to-end accuracy, referred to as ``$\Delta$ w/ Interaction'' in \citet{elgohary-etal-2021-nl}. 

Specifically, future work should include Execution Accuracy (EX\%) along with Exact Set Match (EM\%). As the set of errors made by modern parsers increasingly drifts towards more difficult gold SQL parses, it becomes more likely that the EM\% and EX\% scores will be disjoint. Examining the errors by T5-large, it was common for a gold parse to be expressed with an ``\verb|EXCEPT SELECT|'' clause, whereas the predicted SQL executed identically with a ``\verb|NOT IN|'' clause.

Additionally, as depicted in Table \ref{all_test_sets}, the EX\% score is higher than EM\% for all test sets except for TaBERT. This is due to the fact that TaBERT does not predict values. Instead, it uses the placeholder ``value'' instead of string values, and ``LIMIT 0'' in limit clauses\footnote{We find this odd, as the feedback provided in the TaBERT test set comments on the values}. Though these instances are not judged as incorrect with EM, they are penalized with EX.
% \subsection{Truncation}
% With the proposed generative concept prediction model, only 2 items in the training set are truncated. Without it and including all features, 743 items are truncated. With only the explanation as a feature, 473 items are truncated.

\section{Conclusion}

We present a new model, DestT5 (\underline{D}ynamic \underline{E}ncoding of \underline{S}chemas using \underline{T5}), which achieves a new state-of-the-art correction accuracy on the interactive parsing dataset \textsc{SPLASH}. By using T5 as a schema prediction model, we display better performance compared to classification-based methods. We validate our results on a new test set for interactive semantic parsing based on a modern parser, and offer recommendations for evaluating future systems.

\section*{Limitations}
As mentioned in Table \ref{all_test_sets}, one limitation of the current study is the small scale of the test sets with modern parsers. We encourage future work to emphasize the development and evaluation on these test sets, specifically those which more closely reflect the current SoTA in text-to-SQL (e.g. T5). Additionally, though we have shown using an auxiliary schema prediction model greatly improves the performance of a text-to-SQL system, the addition of a model for the text-to-SQL task is a limitation given the time and training resources required. 

% \section*{Ethics Statement}

% \section*{Acknowledgements}

% Entries for the entire Anthology, followed by custom entries
\bibliography{custom}
\bibliographystyle{acl_natbib}

\appendix

% \section{Example Appendix}
% \label{sec:appendix}

% This is a section in the appendix.

\end{document}